\documentclass[10pt,twocolumn,letterpaper]{article}

\usepackage{wacv}              

\usepackage{graphicx}
\usepackage{multirow}
\usepackage{amsmath}
\usepackage{amssymb}
\usepackage{booktabs}
\usepackage{algorithm}
\usepackage{algpseudocode}
\usepackage{tabularx}
\usepackage{lipsum}
\usepackage{makecell}
\usepackage{enumitem}
\usepackage{pifont}
\usepackage[pagebackref,breaklinks,colorlinks]{hyperref}
\makeatletter
\def\@fnsymbol#1{\ensuremath{\ifcase#1\or \dagger\or \ddagger\or
   \mathsection\or \mathparagraph\or \|\or **\or \dagger\dagger
   \or \ddagger\ddagger \else\@ctrerr\fi}}
    \makeatother

\usepackage[capitalize]{cleveref}
\crefname{section}{Sec.}{Secs.}
\Crefname{section}{Section}{Sections}
\Crefname{table}{Table}{Tables}
\crefname{table}{Tab.}{Tabs.}

\def\wacvPaperID{2158} 
\def\confName{WACV}
\def\confYear{2025}

\begin{document}

\title{ContextIQ: A Multimodal Expert-Based Video Retrieval System for Contextual Advertising}

\author{Ashutosh Chaubey$^{2*}$\thanks{Work done while employed at Anoki Inc.}, Anoubhav Agarwaal$^{1*}$, Sartaki Sinha Roy$^{1*}$, Aayush Agrawal$^{1*}$, Susmita Ghose$^{1*}$ \\
$^{1}$Anoki Inc. $^{2}$University of Southern California \\
{\tt\small achaubey@usc.edu \{anoubhav,sartaki,aayush,susmita\}@anoki.tv}\\
}


\maketitle
\def\thefootnote{*}\footnotetext{These authors contributed equally to this work}\def\thefootnote{\arabic{footnote}}
\begin{abstract}
\vspace{-2mm}
Contextual advertising serves ads that are aligned to the content that the user is viewing. The rapid growth of video content on social platforms and streaming services, along with privacy concerns, has increased the need for contextual advertising. Placing the right ad in the right context creates a seamless and pleasant ad viewing experience, resulting in higher audience engagement and, ultimately, better ad monetization. From a technology standpoint, effective contextual advertising requires a video retrieval system capable of understanding complex video content at a very granular level. Current text-to-video retrieval models based on joint multimodal training demand large datasets and computational resources, limiting their practicality and lacking the key functionalities required for ad ecosystem integration. We introduce \textbf{ContextIQ}, a multimodal expert-based video retrieval system designed specifically for contextual advertising. ContextIQ utilizes modality-specific experts—video, audio, transcript (captions), and metadata such as objects, actions, emotion, etc.—to create semantically rich video representations. We show that our system, without joint training, achieves better or comparable results to state-of-the-art models and commercial solutions on multiple text-to-video retrieval benchmarks. Our ablation studies highlight the benefits of leveraging multiple modalities for enhanced video retrieval accuracy instead of using a vision-language model alone. Furthermore, we show how video retrieval systems such as ContextIQ can be used for contextual advertising in an ad ecosystem while also addressing concerns related to brand safety and filtering inappropriate content.
\end{abstract}

\vspace{-4mm}
\section{Introduction}
\label{sec:intro}
Contextual advertising involves placing ads based on the content a consumer is viewing, improving user experience and leading to higher engagement and ad monetization \cite{Zhang2012ContextualAdvertising}. This method mitigates the need for personal data in advertising, which raises notable legal and ethical concerns \cite{cookie1, cookie2}. Contextual advertising is already widely integrated into ads served on websites powered by platforms like Google AdSense \cite{google_adsense}. Recent AI advancements have taken this a step further by enabling a deeper semantic understanding of multimedia content \cite{aiads}, allowing for more precise ad placements beyond traditional targeting \cite{keyword}. Building on these developments, this study expands the use of contextual advertising to video content, aiming to improve ad targeting on social platforms such as YouTube and streaming services like Free Ad-Supported Streaming Television (FAST) and Video on Demand(VoD). 

Modern video streaming platforms, with their vast content libraries, require advanced methods for analyzing video content to retrieve the most suitable content for ad placements for contextual advertising. We propose that this task of identifying the most relevant video can be achieved with text prompts that align with an advertisement campaign and hence can be formulated as the text-to-video (T2V) retrieval problem in multimodal learning \cite{expert1, clip1, vtr2, vtr8}. Therefore, improvements in T2V retrieval models can enhance contextual advertising by better-aligning ads with the semantic context of video content.

With the exponential growth of video content, T2V retrieval has become more essential, driving progress in search algorithms \cite{search1, search2, search3} and video representation techniques such as masked reconstruction \cite{masked1, masked2, vidmae}, multimodal contrastive learning \cite{chen2023vast, mplug2}, and next-token prediction from video data \cite{nxt1, nxt2}. 

A recent trend in T2V retrieval is large-scale pretraining to learn a joint multimodal representation of a video \cite{wang2024internvideo2scalingfoundationmodels, chen2023vast, valor}. While these joint models have demonstrated strong performance on public benchmarks, their reliance on massive multimodal datasets and substantial computational resources limits their practical use. In contrast, there are expert-based approaches that use specialized models to extract features from individual video modalities such as visuals, motion, speech, audio, OCR, etc. \cite{expert1, expert2}. Though this method is well-established in T2V retrieval, its prominence has declined with the rise of joint multimodal models. 
However, expert models present unique benefits for contextual advertising, which we explore in this work.

This paper introduces ContextIQ, a multimodal expert-based video retrieval system for contextual advertising. ContextIQ leverages experts across multiple modalities—video, audio, captions (transcripts), and metadata such as objects, actions, emotions, etc.—to create semantically rich video representations (ref. Sec. \ref{sec:approach} \& \ref{sec:detail}). Its modular design makes it highly flexible to various brand targeting needs.  For instance, for a beauty brand, we could integrate advanced object detection to spot niche beauty accessories. Moreover, the proposed system can selectively use only a relevant expert model, focusing on the key metadata to enable efficient targeting, making it well-suited for a fast ad-serving system. Furthermore, our system enhances interpretability by allowing individual experts to be analyzed for error tracing and model improvements, which is vital for explaining business outcomes in ad targeting. Beyond conventional video retrieval, we show how ContextIQ can be integrated seamlessly into the ad ecosystem, processing long-form streaming content and implementing brand safety filters to ensure ads are placed within contextually appropriate and safe content (ref. Sec. \ref{sec:contextual_advertising}).

We evaluate the effectiveness of our ContextIQ system across several video retrieval benchmarks, such as MSR-VTT \cite{msrvtt}, Condensed Movies \cite{bain2020condensedmovies}, and a newly curated dataset (\emph{Val-1}) of high-production content that we make public. We compare it to state-of-the-art joint multimodal models \cite{languagebind, clip, onepeace} and commercial solutions like Google’s multimodal embedding API \cite{vertex} and Twelvelabs Marengo \cite{twelvelabs}. The results show that our expert-based approach is better or comparable to these solutions (ref. Sec. \ref{subsec:zero_shot_results}). 

We also address key evaluation challenges, particularly the differences between video domains in existing public datasets—such as shorter, amateur-produced content—and the complex, high-production content typical of contextual advertising by validating the proposed technique on the curated dataset (\emph{Val-1}) of movie contents and evaluating different approaches by manual validators. To further concretize our design choice of having multiple experts, we perform ablation studies on the impact of incorporating multiple modalities on an internal dataset (\emph{Val-2}), demonstrating how our system effectively aggregates modality-specific experts. By leveraging the complementarity of different modalities, ContextIQ achieves improvements in both performance and coverage (ref Sec. \ref{subsec:ablation}). We release our datasets, retrieval queries and annotations publicly at \href{https://github.com/AnokiAI/ContextIQ-Paper}{https://github.com/AnokiAI/ContextIQ-Paper}. In summary, our contributions are as follows, 

\begin{enumerate}[label={(\roman*)},leftmargin=*]
    \vspace{-0.2cm}\item We present ContextIQ, a multimodal expert-based video retrieval system specifically designed for contextual advertising. With capabilities for long-form content processing, modularity for real-time ad serving, and robust brand safety measures, it integrates seamlessly into the advertising ecosystem, effectively bridging the gap between video retrieval research and contextual video advertising.
    \vspace{-0.2cm}\item On existing benchmarks, we show that combining modality-specific experts can yield better or comparable results with state-of-the-art joint multimodal models without the need for joint training, large-scale multimodal datasets, or significant computational resources.
    \vspace{-0.2cm}\item We release a validation dataset (\emph{Val-1}) of high-production content for text-to-video retrieval for contextual advertising and show the superior performance of the proposed approach on that dataset.
    \vspace{-0.2cm}\item We perform ablation studies to show that additional modalities via expert-based models increase the coverage and accuracy of text-to-video retrieval for contextual advertising by showing results on an internal dataset of video contents.
\end{enumerate}

\begin{figure*}
     \centering
     \includegraphics[width=0.75\textwidth]{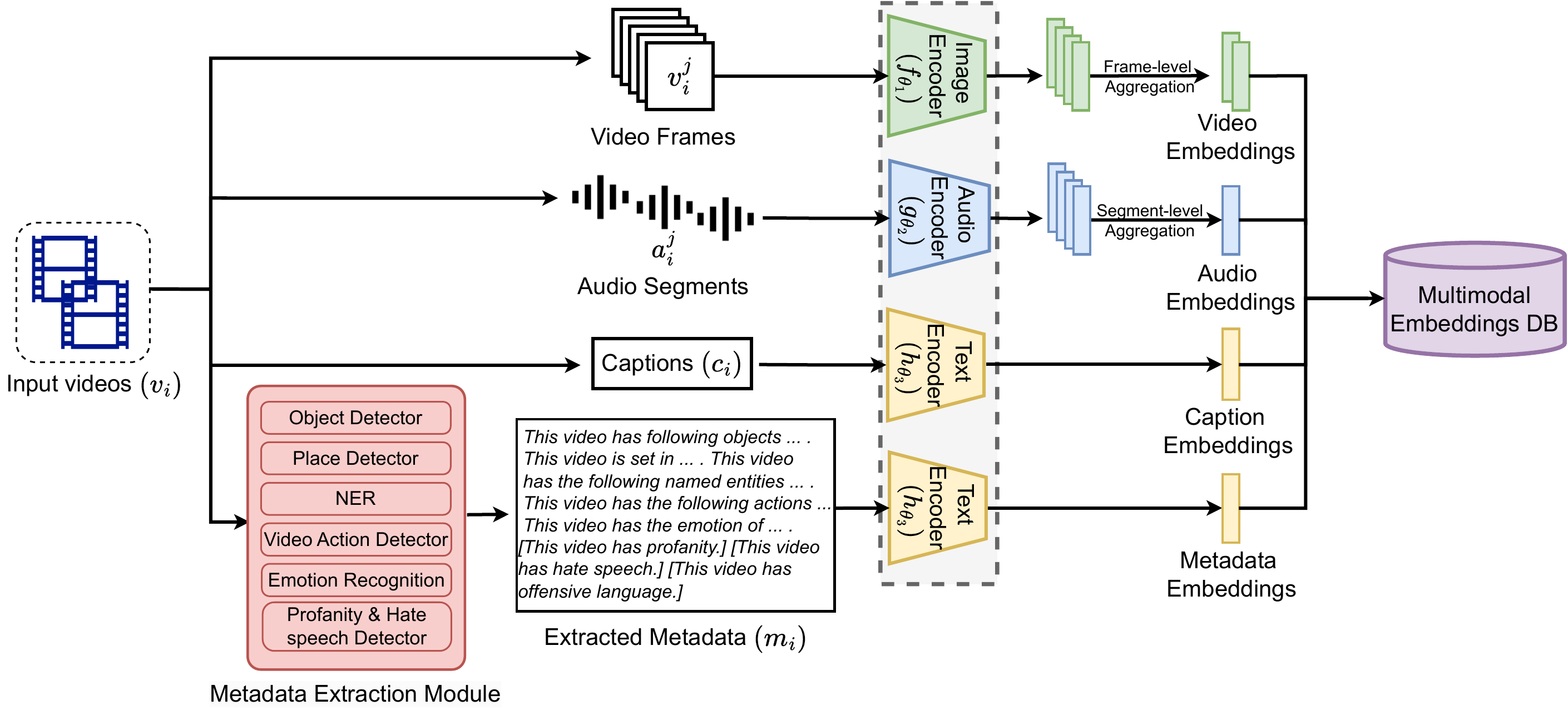}
     
     \vspace{-4mm}
     \caption{Multimodal embedding generation pipeline for ContextIQ. Input videos are processed by the metadata extracting module, which uses expert models for extracting objects, actions, places, etc., and converts it into a metadata sentence ($m_i$). Four modality encoders then encode the video frames, audio, caption (transcripts) and metadata in parallel and store the embeddings in a multimodal embeddings DB.}
     \vspace{-4mm}
     \label{fig:feature_extraction}
 \end{figure*}

\section{Related Works}
\noindent\textbf{AI in advertising.} Liu \etal  \cite{aiads1} employed text-based topic modeling to tag content, enabling buyers to make more informed ad bids. Wen \etal \cite{aiads2} utilized decision trees to identify key attributes that enhance ad persuasiveness, facilitating the selection of suitable ads and insertion points within videos. Bushi \etal \cite{aiads3} analyzed sentiment to prevent negative ad placements, while Vedula \etal \cite{aiads4} predicted ad effectiveness by training separate neural network models on video, audio, and text modalities.

\noindent\textbf{Text-to-Video Retrieval.} Earlier approaches relied on convolutional neural networks (CNNs) and recurrent neural networks (RNNs) \cite{vtr7, vtr8} to extract spatio-temporal features from video frames. However, the advent of Transformer-based architectures \cite{vtr12, vtr13, vtr14} has led to a paradigm shift, with Transformers outperforming traditional CNNs and RNNs in many public benchmark datasets.

\noindent\textbf{Multimodal Pre-training and Joint Learning.} The impact of large-scale pre-training, especially using models like CLIP \cite{clip} that align images with text, has been groundbreaking in the video retrieval domain \cite{clip1, clip2, clip3, clip4, xclip}. For instance, CLIP4Clip \cite{clip1} builds upon CLIP to enable end-to-end video-text retrieval via video-sentence contrastive learning. Similarly, CLAP \cite{laionclap2023} aligns audio with text using contrastive learning. Incorporating additional modalities contrastively can significantly enhance the model's performance and robustness \cite{chen2023vast, wang2024internvideo2scalingfoundationmodels}. LanguageBind contrastively binds language across multiple modalities within a common embedding space \cite{languagebind}. VALOR utilized separate encoders for video, audio, and text to develop a joint visual-audio-text representation \cite{valor}. We show that compared to these jointly trained approaches, ContextIQ performs on par on video-retrieval benchmarks such as MSR-VTT \cite{msrvtt} without any joint training.

\noindent\textbf{Multimodal Experts.}  There have been several works in video retrieval that employ multiple experts to extract features from different video modalities, including scene visuals, motion, speech, audio, OCR, and facial features \cite{expert1, expert2, experts5}. 
This approach introduces the challenge of effectively aggregating these expert-derived features. For example, Miech \etal \cite{expert2} utilizes precomputed features from experts in text-to-video retrieval, where the overall similarity is calculated as a weighted sum of each expert’s similarity score. Liu \etal \cite{expert1} extends the mixture of experts model by employing a collaborative gating mechanism, which modulates each expert feature in relation to the others. \cite{expert3} integrates a multimodal transformer to encode visual, audio, and speech features from different experts, with BERT handling the text query. In contrast to these techniques, we show how our expert-based retrieval system is specifically designed for contextual advertising by providing modularity for targeting flexibility and interpretability, brand safety filters, and ad ecosystem integration.

\section{Approach}
\label{sec:approach}
For a video dataset  $V = \{v_1, ..., v_N\}$, our goal is to develop a text-to-video retrieval system. Conceptually, we aim to learn a similarity function  $\mathcal{S}(v_i, t)$ that assigns higher scores to videos that are more relevant to the given text query $t$. 

We extract embeddings from multiple modalities, including raw video, audio, captions (speech transcripts), and combined visual (objects, places, actions, emotion) and textual (NER, emotion, profanity, hate speech) metadata, as shown in the Metadata Extraction Module in Fig. \ref{fig:feature_extraction}. Since the expert encoders for each modality are not jointly trained, their embeddings reside in separate semantic spaces. These embeddings are then stored in a multimodal embeddings database. We propose a retrieval approach (Fig. \ref{fig:search}) that compares, merges, and re-ranks these modality-specific embeddings to deliver the most contextually relevant videos.

\subsection{Multimodal Embedding Generation}
\label{subsec:multimodal_embedding_generation}
\subsubsection{Video}
\label{subsubsec:video_embedding}
As illustrated in Fig. \ref{fig:feature_extraction}, we consider each video $v_i$ to be composed of $M$ frames $v_i^j$, which are encoded using an image-encoder $f_{\theta_1}$ from a vision-text model \cite{blip2}. This encoding captures intrinsic video features learned by the vision-text model.

We found that splitting the video into fixed non-overlapping time segments, each containing T frames, enhances both retrieval accuracy and localization. For each temporal segment  $C_i^k = \{v_i^z \text{ }|\text{ } z \in \{kT+1, \dots, (k+1)T\}\}$, segment-level embeddings are obtained by applying an aggregation function $\mathcal{A}_v$ to the frame-level features,

\vspace{-0.1in}
\begin{equation}
    \mathbf{e}_i^k = \mathcal{A}_v(\{f_{\theta_1}(v_i^z)\text{ }|\text{ }z \in \{kT+1,\dots, (k+1)T\}\})
\end{equation}
\vspace{-0.1in}

The final set of video-level embeddings for the video $v_i$ that we store is,

\vspace{-0.1in}
\begin{equation}
    \mathcal{F}_i^v = \{\mathbf{e}_i^1, \dots, \mathbf{e}_i^{(M/T)}\}
\end{equation}

\begin{figure*}
     \centering
     \includegraphics[width=0.75\textwidth]{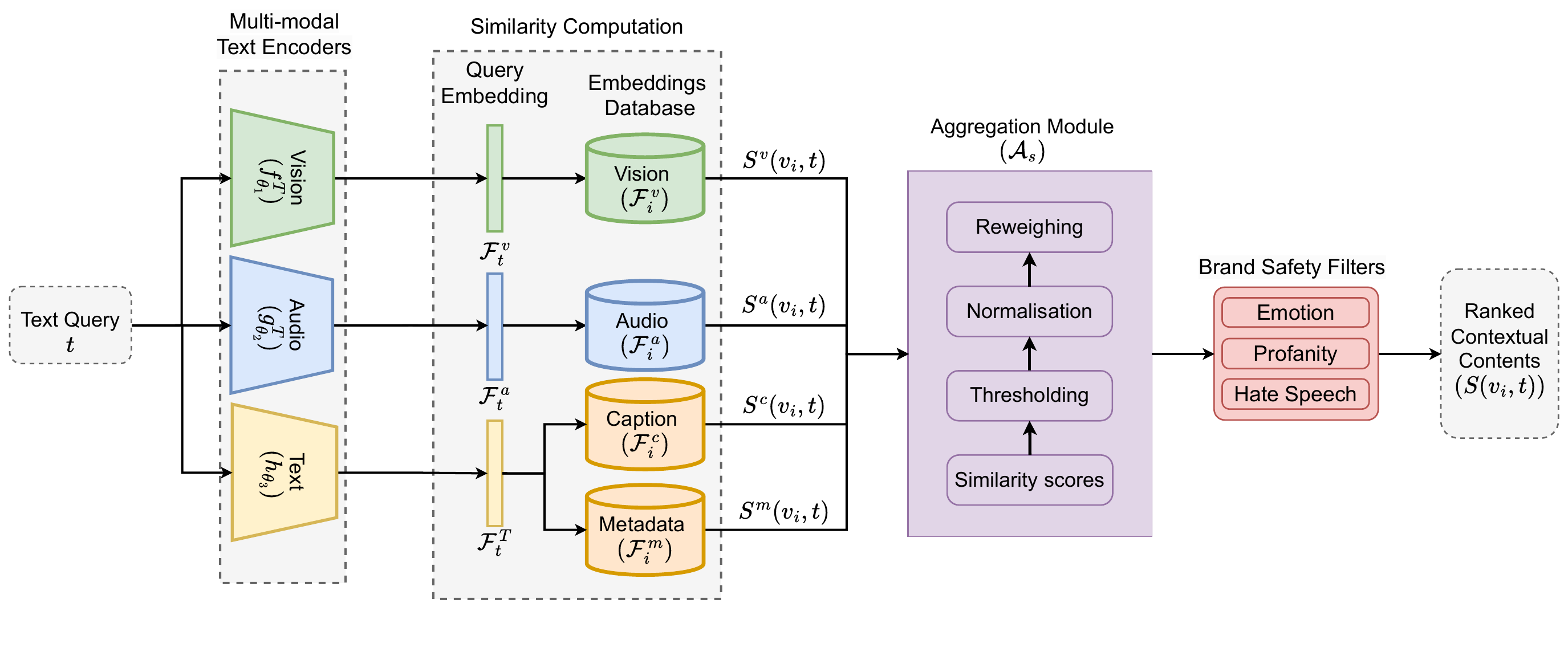}
     \vspace{-7mm}
     \caption{Multimodal search pipeline for ContextIQ. Text query $t$ is first encoded by different text encoders and the multimodal embedding DB is searched to find similar videos. The aggregation module combines the results obtained from different modalities, and the final results are obtained after applying brand-safety filters utilizing emotion, profanity, and hate speech.}
     \vspace{-5mm}
     \label{fig:search}
 \end{figure*}

Note that for a single video, we store $(M/T)$ embeddings in the multimodal database corresponding to intrinsic video features. This ensures that for long-form video content, the features have high temporal resolution and do not average out, resulting in better retrieval.
\vspace{-0.2in}

\subsubsection{Audio}
To capture the non-verbal audio elements in a video, such as sound effects, music, and ambient noise, we utilize the audio encoder $g_{\theta_2}$from an audio-text model \cite{laionclap2023}. While speech and textual content are encoded through captions (ref. Sec. \ref{subsubsec:transcript}), this approach specifically targets the rich auditory features beyond spoken language.

The audio track $a_i$ for a particular video $v_i$ is divided temporally into equal-sized segments $a_i^k$, each of which is encoded using $g_{\theta_2}$. The final audio embedding $\mathcal{F}_i^a$ for the video is obtained by applying an aggregation function $\mathcal{A}_a$ to the features across all the segments:

\vspace{-0.1in}
\begin{equation}
    \mathcal{F}_i^a = \mathcal{A}_a\left(g_{\theta_2}(a_i^1), g_{\theta_2}(a_i^2), \dots, g_{\theta_2}(a_i^K)\right)
\end{equation}
\vspace{-0.3in}

\subsubsection{Caption}
\label{subsubsec:transcript}
To incorporate textual and speech information from a video, we encode the caption (transcript) $c_i$ for a given video using a text encoder $h_{\theta_3}$, resulting in the caption feature embedding of the video:
\vspace{-0.1in}
\begin{equation}
    \mathcal{F}_i^c = h_{\theta_3}(c_i)
\end{equation}
\vspace{-0.3in}

\subsubsection{Metadata}
\label{subsubsec:metadata_ectraction}
Foundation vision models \cite{clip, blip2} excel at learning features for vision-specific tasks with minimal fine-tuning. To enhance the performance of our text-to-video retrieval system, we incorporate models specifically trained for vision and textual tasks, effectively augmenting foundational models (ref. Sec. \ref{subsec:ablation}).

After inference with the task-specific models, we store the extracted information into a metadata sentence $m_i$. This sentence captures objects, places, actions, emotions, named entities and flags any profanity, hate speech, or offensive language. This metadata is vital for contextual advertising. For example, object detection can allow fashion brands to target content featuring clothing or accessories, while place detection can benefit outdoor gear brands by identifying natural landscapes. Video action recognition can enable fitness brands to reach workout-related content, and named entity recognition might help travel agencies focus on videos mentioning tourist destinations. $m_i$ provides a unique and flexible method for integrating outputs from the different expert models. $m_i$ follows the template, as illustrated in Fig. \ref{fig:feature_extraction}.

Finally, we encode $m_i$ by the same text encoder model $h_{\theta_3}$ from Sec. \ref{subsubsec:transcript}, to obtain the metadata feature embedding for the video:

\vspace{-0.15in}
\begin{equation}
    \mathcal{F}_i^m = h_{\theta_3}(m_i)
\end{equation}
\vspace{-0.2in}

Additionally, the emotion, profanity, and hate speech metadata are used as a filtering mechanism during the search process to ensure the retrieval of brand-safe videos. This capability extends beyond the conventional scope of video-text retrieval research, with its application further elaborated in Sec. \ref{sec:contextual_advertising} on contextual advertising.

\subsection{Multimodal Search}
\label{subsec:multimodal_search}



Our system employs multimodal embeddings from audio, video, text and metadata to enable efficient and precise content retrieval. As shown in the Fig. \ref{fig:search}, a text query $t$ is first encoded using the text encoders of the vision-text model ($f_{\theta_1}^T$), audio-text model ($g_{\theta_2}^T$) and text-text model ($h_{\theta_3}$) resulting in the embeddings $\mathcal{F}_{t}^{v}$, $\mathcal{F}_{t}^a$ and $\mathcal{F}_{t}^T$ respectively.

The text embeddings are then compared against their corresponding embedding databases using cosine similarity. Specifically, $\mathcal{F}_t^a$ is compared with audio embeddings $\{\mathcal{F}_i^a\}$, producing similarity scores $\{\mathcal{S}^{a}(v_i, t)\}$; $\mathcal{F}_t^v$ is compared with visual embeddings $\{\mathcal{F}_i^v\}$, yielding similarity scores $\{\mathcal{S}^{v}(v_i, t)\}$; and $\mathcal{F}_t^T$  is compared with metadata embeddings $\{\mathcal{F}_i^m\}$ and caption embeddings $\{\mathcal{F}_i^c\}$ , generating similarity scores $\{\mathcal{S}^{m}(v_i, t)\}$ and $\{\mathcal{S}^{c}(v_i, t)\}$, respectively. Note that $\{\mathcal{S}^{v}(v_i, t)\}$ is the max-similarity score of $t$ with the segments ($\{C_i^1,\dots,C_i^{(M/T)}$) of video $v_i$.

These similarity scores $\{(\mathcal{S}^c, \mathcal{S}^m, \mathcal{S}^a, \mathcal{S}^v)(v_i, t)\}$ are subsequently aggregated into a combined score $\mathcal{S}(v_i, t)$ using an aggregation module $\mathcal{A}_S$ (ref. Fig. \ref{fig:search}) which involves normalization, thresholding, and weighted merging,

\begin{itemize}[leftmargin=*]
    \vspace{-0.2cm}\item \textbf{Normalization.} Scores are standardized within their respective modality space and weighted to produce normalized score:

    \vspace{-0.25in}
    \begin{equation}
        \mathcal{N}^k(v_i, t) = \lambda^k \cdot \frac{\mathcal{S}^k(v_i, t) - \mu^k}{\sigma^k}
    \end{equation}
    \vspace{-0.2in}
    
    where $\mu^k = \frac{1}{n} \sum_i^n \mathcal{S}^k(v_i, t)$ is the mean, $\sigma^k = \sqrt{\frac{1}{n} \sum_{1}^{n} \left( S^k(v_i, t) - \mu^k \right)^2}$ is the standard deviation and $\lambda^k$ is the weight of modality $k$ (metadata, caption, video or audio).

    \vspace{-0.2cm} \item \textbf{Thresholding.}  Scores are thresholded to obtain dictionaries for each modality
    $\mathcal{X}^k = \{(i : \mathcal{N}^k(v_i, t))\text{  }|\text{  } \mathcal{S}^k(v_i, t) > \alpha^k\}$
    where $\alpha^k$ is the modality specific threshold.
    
    \vspace{-0.2cm}\item \textbf{Merging.} The thresholded dictionaries for each modality are merged using a max-aggregation approach, with the final scores $\mathcal{S}(v_i, t)$ determined by the maximum value for each key.

\end{itemize}

The thresholds $\alpha^k$ and weights $\lambda^k$ can be tuned for optimal performance on a representative set of queries. The resulting video contents are finally filtered through the brand safety mechanism (ref. Sec. \ref{sec:contextual_advertising}). 

\section{Implementation Details}
\label{sec:detail}

\subsection{Multimodal encoders}

We use PyTorch \cite{Ansel_PyTorch_2_Faster_2024} for all our model implementations, and we use four NVIDIA RTX 4090 GPUs to run all our experiments.  The text encoder $h_{\theta_3}$ is a pre-trained MPNet model \cite{song2020mpnet} fine-tuned on a set of 1 billion text-text pairs as described by Reimers et al \cite{reimers-2019-sentence-bert}. The vision-text model $f_{\theta_1}$ is a pre-trained BLIP2 Qformer \cite{blip2}, and we use the implementation provided by LAVIS\cite{li-etal-2023-lavis}. We split the video into equal segments of 15 seconds each and sample one out of ten frames for embedding generation (ref. Sec. \ref{subsubsec:video_embedding}). The audio encoder $g_{\theta_2}$ is the CLAP \cite{laionclap2023} model as implemented in \cite{githubGitHubLAIONAICLAP}. Since CLAP is trained on 5-second audio segments, we divide the audio into 5-second chunks ($a_i^k$) for inference. The aggregation functions $\mathcal{A}_a$ and $\mathcal{A}_v$ used during audio and video embedding generation (Sec. \ref{subsec:multimodal_embedding_generation}) are temporal mean pooling functions, shown to be effective over other temporal aggregation techniques \cite{clip1,bain2022cliphitchhikersguidelongvideo}. 

\subsection{Metadata Extraction}
\label{subsec:metadata_extraction}
 We use the YOLOv5 model \cite{ultralytics2021yolov5}, trained on the Objects365 dataset \cite{Shao_2019_ICCV_objects365}, to detect objects within videos with an IOU threshold of 0.45 and a confidence threshold of 0.35. An object is considered present in the video only if it appears in at least 20\% of the frames, to filter false positives. 
 
 For place detection, we finetune a ResNet50 model \cite{He2015DeepRLresnet} on the Places365 dataset \cite{zhou2017places}. Only frames where object detection for the \emph{Person} class covers less than 10\% of the area are used to ensure a clear background. Top predictions from these frames with softmax scores above 0.3 are considered, and the most frequent place prediction is tagged as the video's location. Since video segments are short, we assume a single location per segment. For both place and object detection, predictions are sampled from every 10th frame.

We use a fine-tuned video masked autoencoder model (VideoMAE2) \cite{Wang2023} on the Kinetics 710 dataset \cite{Kay2017, Carreira2018, Carreira2019} for video action recognition. The Kinetics dataset comprises shorter, simpler YouTube clips featuring single actions, whereas our video retrieval algorithm is applied to videos with longer, more complex scenes involving multiple actions. To bridge this gap, we reduced the Kinetics 710 classes to 185 by eliminating less relevant or overly specific classes for advertising, merging correlated classes, and discarding those with low Kinetics validation accuracy. We also refined the majority voting method by incorporating prediction probability scores, improving the handling of multiple actions in a single clip. More implementation details are present in the supplementary material, and we share the list of reduced classes on our github repository - \href{https://github.com/AnokiAI/ContextIQ-Paper}{https://github.com/AnokiAI/ContextIQ-Paper}.

For named entity recognition, we utilize the text-based RoBERTa model, en\_core\_web\_trf from Spacy \cite{Honnibal_spaCy_Industrial-strength_Natural_2020} to extract the named entities from the captions of each video scene. For profanity detection, we used the \emph{alt-profanity-check} \cite{githubGitHubDimitrismistriotisaltprofanitycheck} python package on the video transcript (caption). Additionally, we use a predefined list of words given in \cite{githubGitHubSurgeaiprofanity} to further filter out profane videos.

For hate speech detection, we use a weighted ensemble of two models. First, we use a pre-trained LLAMA 3 8B Instruct model \cite{llama3modelcard} with a temperature of 0.6, leveraging advanced prompting strategies such as JSON-parseable responses and chain-of-thought reasoning to flag content as hateful. Secondly, we use a pre-trained BERT classifier \cite{kim-etal-2022-hate} trained on the HateXplain dataset \cite{mathew2021hatexplain} to categorize content into three classes: hate speech, offensive, and normal. For text emotion recognition, first, we use a pre-trained Emoberta-Large model \cite{kim2021emobertaspeakerawareemotionrecognition} from huggingface \cite{huggingfaceTae898emobertalargeHugging} which is trained on the MELD \cite{poria-etal-2019-meld} and IEMOCAP \cite{Busso2008iemocap} datasets. Furthermore, we also leverage the computed video and audio embeddings from Sec. \ref{subsec:multimodal_embedding_generation} for tagging emotions by associating text concepts with different emotions. For example, we tagged the emotion \emph{joy} with text queries like \emph{people smiling} and \emph{people dancing}, and assigned the emotion \emph{joy} to all videos retrieved through the video (vision) modality using these queries. More implementation details of emotion, profanity, and hate speech detection are present in the supplementary material.

        
        

\begin{table}[t]
    \centering
    \resizebox{0.75\columnwidth}{!}{%
    \begin{tabular}{lcccc}
        \hline \hline
        \textbf{Model} & \textbf{P@1} & \textbf{P@5} & \textbf{R@5} & \textbf{MAP@5} \\
        \midrule
        
        Vertex API\cite{vertex} & \underline{81.9} & 57.0 & 93.2 & 83.1 \\
        LanguageBind\cite{languagebind} & \textbf{85.5} & \textbf{66.6} & \textbf{97.7} & \textbf{86.6} \\
        \textbf{ContextIQ (Ours)} & 81.7& \underline{59.1} & \underline{93.7} & \underline{83.2} \\
        
        \hline \hline
    \end{tabular}
    }
    \vspace{-2mm}
    \caption{Performance comparison on MSR-VTT for ContextIQ, Google Vertex \cite{vertex}  and LanguageBind \cite{languagebind}.}
    \vspace{-3mm}
    \label{tab:msrvtt}
\end{table}

        

\begin{table}[t]
    \centering
    \resizebox{0.75\columnwidth}{!}{%
    \begin{tabular}{lcccc}
        \hline \hline
        \textbf{Model} & \textbf{P@1} & \textbf{P@5} & \textbf{R@5} & \textbf{MAP@5} \\
        \midrule
        
        {TwelveLabs \cite{twelvelabs}} & 96.6 & \textbf{90.3} & 100 & \textbf{95.6} \\
        LanguageBind \cite{languagebind} & 89.7 & 83.5 & 100 & 91.5 \\
        \textbf{ContextIQ (Ours)} & \textbf{96.6} & \underline{88.3} & \textbf{100} & \underline{94.4} \\
        \hline \hline
    \end{tabular}
    }
    \vspace{-2mm}
    \caption{Performance comparison on Condensed Movies \cite{bain2020condensedmovies} for ContextIQ, TwelveLabs \cite{twelvelabs} and LanguageBind \cite{languagebind}.}
    \vspace{-6mm}
    \label{tab:condensed}
\end{table}

\subsection{Validation datasets and Metrics}
\label{subsec:validation_datasets}
\noindent\textbf{Validation datasets.} As mentioned before, we show the efficacy of the proposed approach in comparison to state-of-the-art video retrieval methods on two public datasets, MSR-VTT \cite{msrvtt} and Condensed Movies \cite{bain2020condensedmovies}. We use the 1kA subset of the MSR-VTT test set for evaluation, which consists of 1k videos and 20 text descriptions for each of them. During our analysis, we observed duplicate text descriptions both within individual video clips and across different clips. Hence to assess performance, we randomly sample one caption per video clip. 

Since the MSR-VTT dataset includes a variety of video rather than entertainment-focused content, we also utilize the Condensed Movies dataset \cite{bain2020condensedmovies}, which consists of scene clips from 3K+ movies. For evaluation, we randomly sample 600 scene clips and extract the first minute of each. Based on the movie and scene descriptions, we use ChatGPT \cite{openai2024chatgpt} to generate a set of 29 text queries, focusing on concepts such as objects, locations, emotion, and other contextual elements to search across the 600 video clips. Because the condensed movies dataset is not tagged with the set of queries we obtained, we manually validated the results on this dataset. (ref. Sec. \ref{subsec:zero_shot_results}). 

Furthermore, to show that our system performs well for contextual advertisement targeting, we manually collected a set of 500 movie clips of different genres from YouTube corresponding to different potential advertisement categories. We call this dataset \emph{Val-1}. The selected clips represent one or more of the following `concepts' - \emph{burger}, \emph{concert}, \emph{cooking}, \emph{cowboys and western}, \emph{dog}, \emph{space shuttle}, \emph{sports}, and \emph{army}. Each of the collected movie clips is then annotated with one or more of these concepts by at least 2 annotators, and we take the union of annotated concepts as ground truth. 

We release all the details about the datasets, including text queries used for validation and annotations on GitHub - \href{https://github.com/AnokiAI/ContextIQ-Paper}{https://github.com/AnokiAI/ContextIQ-Paper}.

\noindent\textbf{Metrics.} For all our experiments, we report one or more of the following metrics (i) Precision@K ($P@K$), which is the proportion of retrieved videos marked as correct out of the top $K$ retrieved videos, averaged across all text queries, (ii) Recall@K ($R@K$), which is the average number of queries for which at least one of the top K retrieved results is marked as correct, and (iii) Mean Average Precision@K($MAP@K$): The mean of the average precision scores from 1 to $K$ ($\{1,\dots,K\}$), computed across all queries.

\begin{figure}
    \vspace{-4mm}
    \centering
    \includegraphics[width=0.8\linewidth]{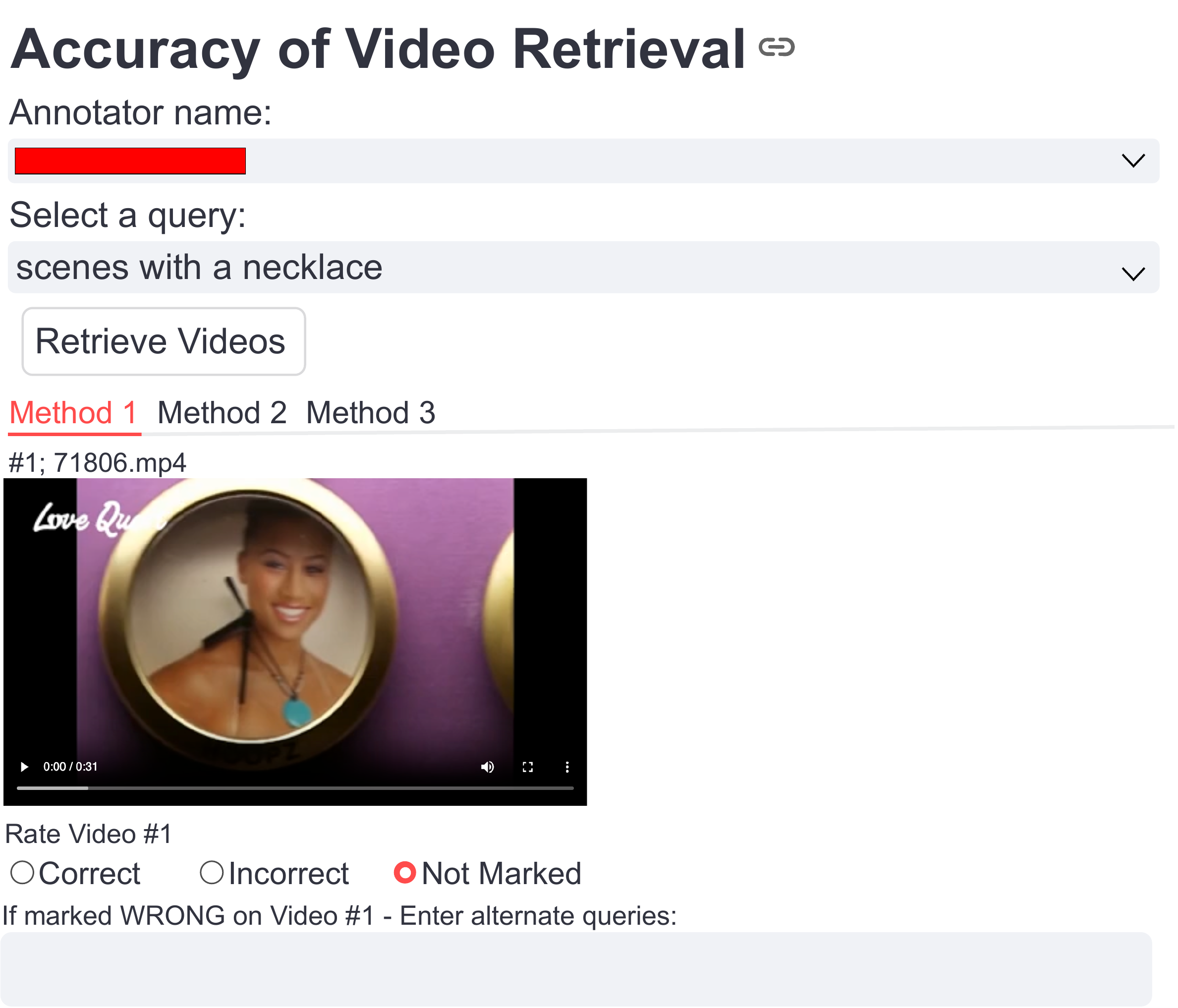} 
    \caption{Validation tool built with Streamlit \cite{streamlitStreamlitFaster}. Note that the different methods are kept anonymous to remove any bias.}
    \vspace{-5mm}
    \label{fig:annotation_ui}
\end{figure}

\section{Results}

\begin{table*}[!t]
    \centering
    \resizebox{0.8\textwidth}{!}{%
    \begin{tabular}{lcccccccccccc}
        \hline \hline
        \textbf{Model}         & \makecell{\textbf{Jointly}\\\textbf{trained}} & \textbf{Modalities} & \textbf{P@5} & \textbf{P@10} & \textbf{P@15} & \textbf{P@20} & \textbf{P@25} & \textbf{P@30} & \textbf{P@35} & \textbf{P@40} & \textbf{P@45} & \textbf{P@50} \\ \hline
        CLIP (Large) \cite{clip1}  & \ding{55}       & \emph{V, L}     & 100 & 100  & 99.2 & 98.8 & 98.5 & 98.8 & 96.8 & 95   & 93.3 & 90.3 \\

LanguageBind \cite{languagebind} & \ding{51}       & \emph{L, V, A, $\delta$, $I_r$}     & 100 & 98.8 & 98.3 & 98.1 & 98.5 & 98.8 & 97.1 & 95.9 & 94.7 & 92   \\
One-Peace \cite{onepeace}    & \ding{51}       & \emph{L, V, A}     & 92.5 & 91.3 & 92.5 & 94.4 & 93.5 & 93.3 & 93.2 & 92.5 & 90.8 & 90.3 \\
\textbf{ContextIQ (Ours)}     & \ding{55}      & L, V, A     & \textbf{100} & \textbf{100}  & \textbf{100}  & \textbf{99.4} & \textbf{99}   & \textbf{98.8} & \textbf{98.2} & \textbf{98.1} & \textbf{97.2} & \textbf{97.3} \\
        \hline \hline
    \end{tabular}
    }
    \caption{Performance comparison on the curated dataset (\emph{Val-1}) for ContextIQ, LanguageBind \cite{languagebind}, One-Peace \cite{onepeace} and CLIP-Large \cite{clip}. \emph{V}: vision, \emph{L}: language, \emph{A}: audio, \emph{$\delta$}: depth, \emph{$I_r$}: infrared.}
    \vspace{-5mm}
    \label{tab:internal}
\end{table*}

\subsection{Zero-shot text-to-video retrieval }
\label{subsec:zero_shot_results}
We evaluate the performance of ContextIQ for text-to-video retrieval using the validation datasets mentioned in Sec. \ref{subsec:validation_datasets}. Tab. \ref{tab:msrvtt} shows the performance of the proposed approach on the MSR-VTT dataset \cite{msrvtt} compared to a state-of-the-art jointly trained multimodal model (LanguageBind \cite{languagebind}) and a popular industry solution for video retrieval (Google's Vertex API \cite{vertex}). For LanguageBind we use the huggingface implementation - \emph{LanguageBind\_Video\_FT} \cite{huggingfaceLanguageBindLanguageBind_Video_FTHugging}. Although ContextIQ is not jointly trained on multiple modalities, it performs slightly better than Google's Vertex. We hypothesize that LanguageBind's superior performance on MSR-VTT can be attributed to its joint training on the large-scale 10M VIDAL dataset, which includes a diverse range of videos similar to the general content found in MSR-VTT, rather than being focused on entertainment-specific content. Additionally, LanguageBind incorporates modalities such as depth, which are absent in ContextIQ.

As shown in Tab. \ref{tab:condensed}, for the Condensed Movies dataset \cite{bain2020condensedmovies}, we compare the results of our proposed technique with \emph{TwelveLabs}, utilizing their \emph{Marengo} API \cite{twelvelabs}, a jointly trained multimodal model for text-to-video retrieval. For all the approaches listed in Tab. \ref{tab:condensed}, we first retrieve videos for each of the text queries generated by ChatGPT (ref. Sec. \ref{subsec:validation_datasets}) and then ask three manual validators to validate the top 5 results using the validation tool built using \emph{Streamlit} \cite{streamlitStreamlitFaster} as shown in Fig. \ref{fig:annotation_ui}. We use a voting-based system among the annotators to compile the results of our validation. We can see that ContextIQ performs better than LanguageBind on all the metrics, while it performs comparable to TwelveLabs. These findings further show that ContextIQ, a mixture of expert-based models can perform comparable to or even better than the jointly trained multimodal models.

Tab. \ref{tab:internal} shows the performance of the proposed approach as compared to different approaches on our curated dataset (\emph{Val-1} as described in Sec. \ref{subsec:validation_datasets}). We use the large variant of CLIP \cite{clip} for our comparison. Note that One-Peace \cite{onepeace} and LanguageBind \cite{languagebind} are multimodal models that are trained jointly, where embeddings from all modalities reside in the same space. ContextIQ performs significantly better than the baselines, especially when we check the precision at higher k values. It is also important to highlight that CLIP, which is just a vision-language model, performs comparably to the multimodal approaches LanguageBind and One-Peace, highlighting that for most of the queries, only visual understanding results in good retrieval results.

\subsection{Ablation: Impact of additional experts and modalities}

\label{subsec:ablation}

We saw in the previous paragraph that only a vision-language model achieves comparable results on the task of video retrieval for advertising on our advertising-focused curated dataset (\emph{Val-1}). In this section, we perform an ablation study to see the efficacy of different modalities when combined with the vision-text model encoder. 

Since we want to simulate the effect of using our system in a large database of multimedia content, we utilize an internal dataset (\emph{Val-2}) comprising over 2,000 long-form videos (movies, TV, and OTT contents) processed through our ContextIQ system (ref. Sec. \ref{sec:contextual_advertising}) to generate over 100,000 scenes (videos), averaging 30 seconds in duration. We curate retrieval query sets for each additional modality, focusing on queries that highlight their individual strengths and are relevant to ad targeting. Details about the dataset are included in our github repository.

We compute the audio, video, caption and metadata embeddings for the entire dataset and store them separately. We then compare the performance of vision-only (video) embeddings with vision combined with different modalities as shown in Tab. \ref{tab:additional}. Since we do not have labeled ground truth for this internal dataset as well, we employ a similar manual validation technique which was used for validating Condensed Movies (ref. Tab. \ref{tab:condensed}). For each query, we search and retrieve the top 30 videos from the dataset. Vision-only results are based on similarity scores between the text query and vision embeddings, while vision + additional modality results combine scores from both modalities using the aggregation module (ref. Sec. \ref{subsec:multimodal_search}). The top 30 videos from both methods are then annotated for correctness by 3 annotators.


\begin{table}[t]
    \centering
    \renewcommand{\arraystretch}{1.1} 
    \resizebox{\columnwidth}{!}{%
    \begin{tabular}{llcccccc}
        \hline \hline
        \textbf{Query set} & \textbf{Modalities} & \textbf{P@5} & \textbf{P@10} & \textbf{P@15} & \textbf{P@20} & \textbf{P@25} & \textbf{P@30} \\
        \hline
        \multirow{2}{*}{\makecell[c]{Metadata set \\ ($\Delta_{\text{avg}} = 4.08$)}} & \emph{V} & 85.7 & 84.3 & 80.5 & 79.5 & 77.6 & 76.2 \\
        & \emph{V} + \emph{M} & \textbf{87.9} & \textbf{86.4} & \textbf{85.0} & \textbf{83.6} & \textbf{83.0} & \textbf{82.4} \\
        \hline
        \multirow{2}{*}{\makecell[c]{Caption set \\ ($\Delta_{\text{avg}} = 5.42$)}} & \emph{V} & 84.2 & 79.5 & 75.4 & 76.6 & 75.4 & 74.6 \\
        & \emph{V} + \emph{L} & \textbf{84.2} & \textbf{82.1} & \textbf{83.5} & \textbf{83.4} & \textbf{82.5} & \textbf{82.5}
 \\
        \hline
        \multirow{2}{*}{\makecell[c]{Audio set \\ ($\Delta_{\text{avg}} = 5.67$)}} & \emph{V} & 85.7 & 82.9 & 79.0 & 83.6 & 83.4 & 84.8 \\
        & \emph{V} + \emph{A} & \textbf{88.6} & \textbf{87.1} & \textbf{87.6} & \textbf{89.3} & \textbf{90.3} & \textbf{90.5}
 \\
        \hline\hline
    \end{tabular}%
    }
    \vspace{-2mm}
    \caption{Performance gain on Adding Modalities to Vision-Only System. \emph{V}: vision, \emph{L}: language (captions), \emph{A}: audio, \emph{M}: metadata}
    \label{tab:additional}
    \vspace{-3mm}
\end{table}

Tab. \ref{tab:additional} shows that adding an additional modality consistently improves precision across all $K$ values compared to using a vision-only model. The precision gap between vision and vision+modality widens as $K$ increases. $\Delta_{avg}$ represents the average difference in precision between vision and vision+additional modality:

\vspace{-0.1in}
    \begin{equation}
        \Delta_{\text{avg}} = \frac{1}{|\mathcal{K}|} \sum_{K \in \mathcal{K}} |P_{V,K} - P_{V+X, K}|
    \end{equation}
    \vspace{-0.15in}
    
Where, $\mathcal{K}$ is the set of $K$ values $\{5, 10,\dots, 30\}$ and $P_{V, K}$ is the precision for vision-only variant at $K$, and $P_{V+X, K}$ is the precision for vision+additional modality.

We observe that the average precision delta is highest for audio, followed by caption and then metadata. This is due to the fact that the vision modality doesn’t capture raw audio or captions (transcripts) but is better at representing metadata elements like objects, actions, places, etc. 

\begin{table}[t]
    \centering
    \resizebox{0.75\columnwidth}{!}{
    \begin{tabular}{lcccccc}
        \hline \hline
        \textbf{Modality} & \multicolumn{6}{c}{\textbf{Overlap \% in Top-K with Vision Modality}} \\
        \cmidrule(l){2-7}
        & @5 & @10 & @15 & @20 & @25 & @30 \\
        \midrule
        Metadata (M) & 2.96 & 5.93 & 6.17 & 6.48 & 7.70 & 7.78 \\
        Caption (L) & 1.11 & 0.56 & 0.74 & 1.39 & 2.22 & 2.22 \\
        Audio (A) & 0.00 & 0.00 & 0.00 & 0.00 & 0.00 & 0.95 \\
        \hline\hline
    \end{tabular}}
    \vspace{-2mm}
    \caption{Overlap percentage in Top-K results for vision only and vision + additional modality.}
    \label{tab:overlap}
    \vspace{-5mm}
\end{table}

 \begin{figure*}
     \centering
     \includegraphics[width=0.75\textwidth]{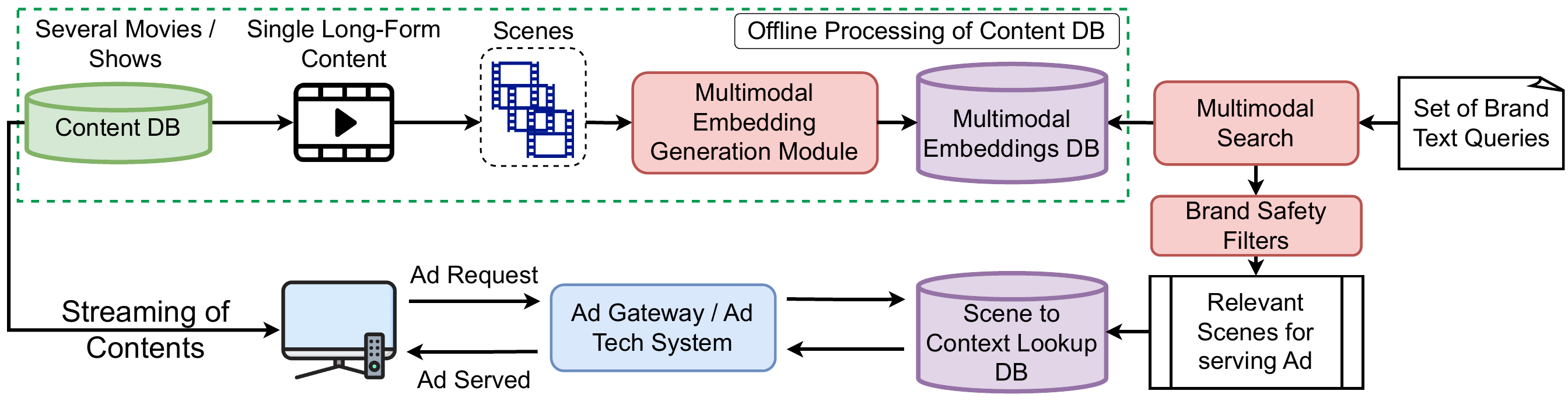}
     \vspace{-3mm}
     \caption{End-to-End ContextIQ video retrieval system for contextual advertising (ref. Sec. \ref{sec:contextual_advertising})}
     \vspace{-5mm}
     \label{fig:contextiq}
 \end{figure*}


Tab. \ref{tab:overlap} further shows that without our aggregation module, the scenes retrieved by each modality independently for the same query differ significantly from those retrieved by the vision-only variant, resulting in very low overlap. This confirms that each modality captures distinct aspects of the video. The overlap with vision follows the order: audio $<$ caption $<$ metadata, which aligns with expectations.
These findings show that utilizing the complementarity of various modalities improves both performance and coverage. 

\section{Contextual Advertising}
\label{sec:contextual_advertising}

Contextual Advertising targets ads based on the content a user is viewing, improving the experience, boosting engagement, and increasing conversion rates, all without using personal information, making it privacy-friendly. Fig. \ref{fig:contextiq} shows how ContextIQ is integrated into the Connected TV advertising ecosystem that enables advertisers to perform contextual advertising. 

\label{longform_content}
\noindent\textbf{Processing Long-Form content.} Long-form content, such as movies and shows, is processed by ContextIQ by breaking them into shorter videos using scene detection. We use PySceneDetect \cite{scenedetectHomePySceneDetect} for scene detection with default parameters. Each scene is subsequently processed through ContextIQ’s multimodal embedding generation module (ref. Sec. \ref{subsec:multimodal_embedding_generation}) to generate the reference multimodal scene embeddings.

\noindent\textbf{Integration into Ad Serving system.} Depending on the brand campaign and the advertisements to be served, an advertiser defines a set of relevant text queries. For example, a pet food brand might have queries such as \emph{dogs}, \emph{cats}, \emph{pet food}, etc. Using these brand-specific text queries and the multimodal embeddings, ContextIQ’s multimodal search (ref. Sec. \ref{subsec:multimodal_search})  identifies scenes where creatives can be contextually served. Additionally, these scenes can be passed through brand safety filters to ensure that brands don't get associated with sensitive/profane scenes. The selected scenes are stored in the scene-to-context lookup DB. When a viewer is watching the TV, the ad gateway looks up the scene to context lookup DB to find out the relevant context for showing advertisements; the ad gateway serves the brand's ad as shown in Fig. \ref{fig:contextiq}. ContextIQ can easily be extended to retrieve relevant scenes for showing an ad based on image, video or even audios. For example, an advertiser might directly use their brand advertisement as a query for video retrieval and get the relevant search results (more details in the Supplementary material). 

\noindent\textbf{Brand Safety.} Ensuring brand safety is crucial in video retrieval systems for contextual advertising. Advertisers are increasingly vigilant about the environments in which their brands appear, as association with inappropriate content, such as offensive language, hate speech, negative emotions, adult material, or references to crime and terrorism, can cause significant reputational damage and erode consumer trust. To address these concerns, our ContextIQ system integrates a safety mechanism comprising two key filters: (i) \textbf{Emotion Recognition} filter, which evaluates the emotional quality of content, ensuring alignment with the brand's messaging. (ii) \textbf{Hate Speech and Profanity Detection} filter, which blocks content containing hate speech, explicit language, or other inappropriate communication, preventing ads from appearing alongside harmful content. Since we already extract emotional and profanity information during metadata extraction (ref. Sec. \ref{subsubsec:metadata_ectraction} \& \ref{subsec:metadata_extraction}), we use the same information during brand safety filtering with no additional compute.

\noindent\textbf{Modularity.} ContextIQ, with its diverse set of expert models, offers flexibility by allowing the use of a specific subset of models tailored to particular use cases. For example, the textual modality can be used for real-time content ingestion to serve ads during live news segments, filtering out violent content that many brands prefer to avoid. Additionally, each expert model can be fine-tuned to meet specific brand requirements. For instance, place and object detection models can be fine-tuned accordingly to support a casino brand looking to detect both a casino location and a roulette wheel within the content.

\section{Conclusion}
This paper introduces ContextIQ, an end-to-end video retrieval system designed for contextual advertising. By leveraging multimodal experts across video, audio, captions, and metadata, ContextIQ effectively aggregates these diverse modalities to create semantically rich video representations. We demonstrate strong performance on multiple video retrieval benchmarks, achieving results better or comparable to jointly trained multimodal models without requiring extensive multimodal datasets and computational resources. Our ablation study shows the advantage of incorporating multiple modalities over a vision-only baseline. We further examine how ContextIQ extends beyond the conventional video retrieval task by integrating seamlessly into the ad ecosystem, processing streamed long-form content, offering modularity for efficient real-time ad serving, and implementing brand safety filters to ensure ads are placed within contextually appropriate and safe content. 

{\small
\bibliographystyle{ieee_fullname}
\bibliography{egbib}
}

\newpage
\appendix
\section{Alternative Modality Queries to ContextIQ}
The flexibility of our system allows us to effortlessly perform queries across different modalities, including video, audio, and image, ensuring any-to-any search capabilities.

\textbf{Image Query:} The process begins by encoding the input query image using the vision encoder of the vision-text model $f_{\theta_1}$, resulting in an image embedding. This embedding is then compared against the vision embeddings of all available content $\{\mathcal{F}_i^v : i = 1,2,...,N\}$ using cosine similarity. The system retrieves and ranks content based on these similarity scores.

\textbf{Video Query:} For video queries, the system first extracts frame-level embeddings from the sampled frames of the query video using the vision encoder of the vision-text model $f_{\theta_1}$. These frame-level embeddings are then aggregated using the previously defined aggregation function $\mathcal{A}_v$ to generate a single video embedding. This video embedding is compared directly with the vision embedding database $\{\mathcal{F}_i^v : i = 1,2,...,N\}$ using cosine similarity. The content is then ranked according to similarity, with the most relevant videos appearing at the top of the results.

\textbf{Audio Query:} The process for audio queries begins by segmenting the query audio into a fixed number of segments. Each segment is encoded using the audio encoder of the audio-text model $g_{\theta_2}$. These segment-level encodings are then aggregated using the previously defined aggregation function $\mathcal{A}_a$ to form a single audio embedding. This aggregated embedding is compared against the audio embeddings database $\{\mathcal{F}_i^a : i = 1,2,...,N\}$ using cosine similarity. The results are then ranked based on these similarity scores.

\section{Video Action Recognition}

\subsection{Simplifying Kinetics 710 classes}
Reducing Kinetics 710 \cite{Kay2017, Carreira2018, Carreira2019} classes to minimize inter-class confusion can be done by either discarding irrelevant classes or combining similar ones. A hierarchical approach to combining Kinetics classes was explored in \cite{hier} using a clustering method. However, this approach only provides examples rather than hierarchical clustering for the entire Kinetics dataset. In our ContextIQ system, we reduced the number of classes by collecting various signals and manually determining which classes to discard or combine. As a result, the number of classes was reduced from 710 to 185. The result is captured in this \href{https://github.com/AnokiAI/ContextIQ-Paper/blob/master/supplementary/video_action_recognition/simply_kinetics710_to_185_classes.xlsx}{sheet} (as referred in the following paragraphs) present in our GitHub repository \href{https://github.com/AnokiAI/ContextIQ-Paper}{https://github.com/AnokiAI/ContextIQ-Paper/}. The signals used were:
\begin{enumerate}

    \item \textbf{Relevance to contextual advertiser}: Some classes, like "stretching arm" or "shuffling feet" may be too mundane, while others, like "playing oboe" or "clam digging," are too niche for a broad audience targeting. Using GPT-4, we identified and marked about 50\% of classes as irrelevant for audience targeting (highlighted in red in the attached sheet). Examples of discarded classes include "Playing oboe" (niche instrument with limited audience), "Pole vault" (niche sport), and "Stretching leg" (too general for segmentation).
    
    GPT4 Prompt: \newline \textit{I have a list of 710 actions. Create a downloadable sheet with three columns, 710 actions, Discard, Reason. The discard should be Yes, but only if it seems less useful for detecting an action. If Discard is yes, also mention the reason (1-2 lines). I want to discard about 50\% of the actions to keep the most useful half. An action is less useful if it does not seem helpful for creating audience segments for ad targeting. E.g, pinching action does not seem useful to target. Do not make any action class. Use the 710 as it is}. \newline\textit{[[Paste the list of 710 actions]]}

    \item \textbf{Groupings and correlated classes}: Many classes in the Kinetics 710 set have high overlap, and their occurrence is highly correlated. For example, there are three separate classes for playing guitar, strumming guitar, and tapping guitar, which the model finds difficult to differentiate, leading to higher inter-class confusion. To address this, we used two approaches: one extends the existing Kinetics 400 \cite{Kay2017} groupings to 710 classes, and the other examines the top correlated classes during prediction. 
    
    \textbf{K400 Groupings}: The K400 set \cite{Kay2017} provided groupings of the 400 classes into 37 groups. However, these groupings were not extended to the additional 300 classes in the K710 set. For these additional 300 classes, we inferred their groupings by finding their text similarity with the 37 groups using the text encoder $h_{\theta_3}$ used in the main paper and tagging the class to the most similar group. These classes are marked with a double asterisk in the K400 grouping column in the sheet.

    \textbf{Top-3 correlated classes}: The VideoMAE2 model \cite{vidmae} generates logits for 710 classes during inference. We build a co-occurrence matrix by counting every pair of classes in the Top-10 logit scores, then compute the correlation matrix. This process is applied to both the Kinetics validation set (50,000 videos) and our internal movie/TV clip set (Sec. 5.2 in the main paper). For instance, classes like dunking, dribbling, shooting, and playing basketball are highly correlated, allowing us to merge them into a single class, such as playing basketball.

    
    \item \textbf{Accuracy on the K710 validation set}: Some classes perform poorly on the simpler Kinetics validation set (likely the same data distribution they were trained on), making them less likely to perform well on our movie clip dataset. We calculate the Top-1 and Top-3 accuracy for each class on the Kinetics set and highlight those in the bottom 25th percentile in the sheet. For instance, photobombing has a 38.8\% Top-1 and 51\% Top-3 accuracy, making it a candidate for discarding to reduce false positives and inter-class confusion.

    \item \textbf{Class occurrence ranking}: Kinetics includes many classes that rarely occur in the wild, such as wood burning (art), stacking die, and wrestling alligator. In our large, diverse internal dataset, we found that 90\% of the Top-3 search results come from only 218 classes \ref{fig:rarety}. In the attached sheet, we list the occurrence count rank of each class (from 1 to 710) and highlight those beyond the top 218.

\begin{figure}[htp]
    \centering
    \includegraphics[width=8cm]{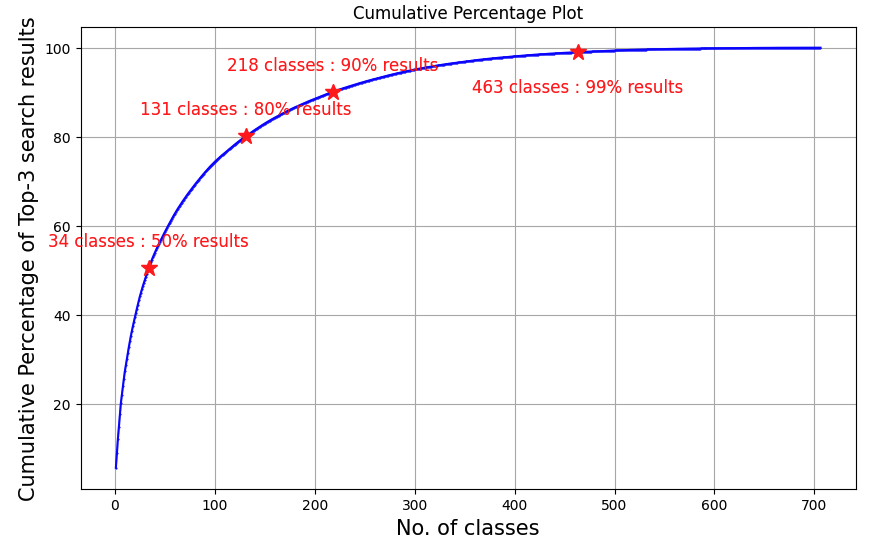}
    \caption{Cummulative Percentage Plot of Top-3 predicted actions on the internal set}
    \label{fig:rarety}
\end{figure}

\end{enumerate}

Using the above four signals, the 710 classes were manually screened and reduced to \textbf{185 classes}:
\begin{itemize}
    \item 182 classes were discarded.
    \item 418 classes were combined into 89 classes (see Tab. \ref{tab:combined_class} for some of the obtained combined classes).
    \item 96 classes were retained.
\end{itemize}

\begin{table}[]
\centering
\caption{Few examples of obtained combined classes}
\label{tab:combined_class}
\begin{tabular}{|c|p{5cm}|}
\hline
Combined class & Actions belonging to the class  \\
\hline
playing cards & playing poker, shuffling cards, card stacking, card throwing, dealing cards, playing blackjack\\ 
drinking alcohol & uncorking champagne, bartending, drinking beer, drinking shots, tasting beer, playing beer pong, pouring beer, tasting wine, pouring wine, opening bottle (not wine), opening wine bottle \\
riding animal & riding camel, riding elephant, riding mule, riding or walking with horse  \\
playing board game &  playing monopoly, playing checkers, playing dominoes, playing mahjong, playing scrabble \\
cleaning floor & cleaning floor, mopping floor, sweeping floor, brushing floor, vacuuming floor, sanding floor \\
\hline
\end{tabular}
\end{table}
Combining classes involves a trade-off between losing specificity and improving average precision @ K and prediction confidence. For instance, while predicting a broader class such as “drinking alcohol” (refer to row 2 of Table \ref{tab:combined_class}) can yield higher precision, it sacrifices the ability to differentiate between specific types like wine and beer.

\section{Emotion Recognition}

\noindent\textbf{Text-based Emotion Recognition.} We use a pre-trained Emoberta-Large \cite{kim2021emobertaspeakerawareemotionrecognition} model, which is trained on the MELD \cite{poria-etal-2019-meld} and IEMOCAP \cite{Busso2008iemocap} datasets, for text-based emotion recognition as it is a speaker-aware model and shows better performance empirically on movie scene subtitles. 

\noindent\textbf{Leveraging Visual and Audio Cues for Emotion Recognition}. The text-based models work only when there is enough text for the model to make a prediction. Moreover, it is difficult to find subtitles for some content, but still, we need to predict emotions in them for better retrieval and brand safe filtering. Since we already use the vision-text model and audio-text models for different parts of the ContextIQ system, we use these models to get some extra signals for predicting emotion. For example, we tagged the emotion \emph{joy} with text queries like \emph{people smiling} and \emph{people dancing}, and assigned the emotion \emph{joy} to all videos retrieved through the video (vision) modality using these queries. Concretely, we associate textual concepts that can be linked to different emotions and then find the scenes that have high video embedding similarity with the emotional text concept. Assume $Q_{t} = \{t:e\}$ to be the text concept dictionary which contains strings $t$ associated with different emotions $e$. Then, for a particular video scene $x$, we say that it is associated to an emotion $e$ if,

\begin{equation}
    f_{\theta_1}(x)\cdot f_{\theta_1}^T(t) > \tau_e
\end{equation}

where $f_{\theta_1}$ and $f_{\theta_1}^T$ are the video and text encoders, respectively, of the vision-text model \cite{blip2}, and $\tau_e$ is a predefined threshold for the concept-emotion pair $t:e$. 

\begin{table*}[]
\centering
\caption{Classification metrics for LLM, BERT and Ensemble Model}
\label{tab:classification_metrics}
\begin{tabular}{l|llll|llll}
          & \multicolumn{4}{c|}{Explicit Hate vs Normal Speech} & \multicolumn{4}{c}{Implicit Hate vs Normal Speech} \\ \hline
Metric &
  LLM &
  BERT &
  \begin{tabular}[c]{@{}l@{}}Ensemble \\ (OR, $\theta$ = 0.7)\end{tabular} &
  \begin{tabular}[c]{@{}l@{}}Ensemble \\ (AND, $\theta$ = 0.2)\end{tabular} &
  LLM &
  BERT &
  \begin{tabular}[c]{@{}l@{}}Ensemble \\ (OR, $\theta$ = 0.7)\end{tabular} &
  \begin{tabular}[c]{@{}l@{}}Ensemble \\ (OR, $\theta$ = 0.2)\end{tabular} \\ \hline
Accuracy  & 83.9        & 77.7        & 81.5       & 85.3       & 75.3        & 63.4       & 73         & 73.2       \\
Precision & 78.9        & 75.9        & 74.5       & 82         & 75.2        & 66.8       & 70.7       & 76.9       \\
Recall    & 92.3        & 81.1        & 95.1       & 90.2       & 74.9        & 52.4       & 78.1       & 66         \\
F1 Score  & 85.1        & 78.4        & 83.5       & 85.9       & 75.1        & 58.8       & 74.2       & 71        
\end{tabular}
\end{table*}

\begin{table*}[]
\centering
\caption{Differential Analysis for different prompting strategies}
\label{tab:diff_analysis}
\begin{tabular}{c|ccccc|ccccc}
                         & \multicolumn{5}{c|}{Explicit Hate vs Normal Speech} & \multicolumn{5}{c}{Implicit Hate vs Normal Speech} \\ \hline
Reasoning                & Yes            & No            & Yes  & Yes  & Yes  & Yes           & No            & Yes  & Yes  & Yes  \\ \hline
Definition of Hate Speech & Yes            & Yes           & Yes  & Yes  & No   & Yes           & Yes           & Yes  & Yes  & No   \\ \hline
Number of Examples       & 3              & 3             & 1    & 0    & 3    & 3             & 3             & 1    & 0    & 3    \\ \hline
Recall                   & 94.6           & \textbf{97.2} & 95.2 & 93.5 & 94.9 & 76.5          & \textbf{85.6} & 74.8 & 77.8 & 75.5 \\
Precision                & \textbf{73.9}  & 65.8          & 72.3 & 70.2 & 71.0 & \textbf{70.3} & 62.9          & 67.3 & 66.3 & 67.4 \\
Accuracy                 & \textbf{80.8}  & 73.4          & 79.4 & 77.1 & 78.7 & \textbf{71.9} & 67.6          & 69.2 & 69.3 & 69.5 \\
F1 Score                 & \textbf{83.0}  & 78.4          & 82.2 & 80.2 & 81.2 & \textbf{73.3} & 72.5          & 70.8 & 71.6 & 71.2
\end{tabular}
\end{table*}

Empirical results show that textual emotion concepts work well only for \textit{joy} emotion. For other emotions, either it is difficult to find emotional text concepts which are relevant to that emotion, or the text concept associated to the emotion is not well represented by the vision-text model.

Similar to visual concepts, we associate audio concepts to different emotions given by $Q_a = {a:e}$, which contains audio files $a$ and corresponding emotion $e$ associated with that audio file. Then for a particular video scene $x$, we say that it is associated to an emotion $e$ if, 

\begin{equation}
    g_{\theta_2}(x_a)\cdot g_{\theta_2}(a) > \tau_e
\end{equation}

where $g_{\theta_2}$ is the audio encoder of CLAP \cite{laionclap2023}, $x_a$ is the audio for the given video and $\tau_e$ is a predefined threshold for the concept-emotion pair $a:e$. Note that we do not use the text encoder of CLAP because text-audio matching did not result into as good results as audio-audio matching. We have only linked audio emotion concepts to \textit{sad} emotion because the rest of the emotions do not show good results empirically.

\section{Hate Speech Detection}


\textbf{Aggregation Strategy}: To combine predictions from the BERT model, the scores for the Hate Speech and Offensive classes are summed. This aggregated score is then compared against a threshold of $\theta = 0.7$. The final prediction is obtained by applying a logical OR operation between the thresholded BERT prediction and the predictions from the LLM to boost recall.

\textbf{Prompting Strategies}:
We implement various prompting techniques to enhance the predictive performance of the LLM \cite{guo2024investigationlargelanguagemodels}. 
\begin{enumerate}[leftmargin=*] 
    \vspace{-0.2cm}\item \textbf{Few-Shot Learning}: A few examples are provided to the model to establish task context, improving its ability to accurately identify hate speech. Specifically we use three examples for the same.
    \vspace{-0.2cm}\item \textbf{Definition of Hate Speech}: A precise definition of hate speech is included in the prompt to ensure consistent detection aligned with the dataset annotations. We use the following definition of hate speech : \emph{Language that disparages a person or group on the basis of protected characteristics like race, gender, and cultural identity.} 
    \vspace{-0.2cm}\item \textbf{Structured JSON Output}: The model is instructed to return its response in JSON format, enabling easy parsing and seamless integration with the contextIQ system. 
    \vspace{-0.2cm}\item \textbf{Chain of Thought Reasoning}: The model is prompted to generate intermediate reasoning steps before determining whether content qualifies as hate speech, enhancing prediction accuracy. \cite{10.5555/3600270.3602070}  

\noindent Various analyses were performed to evaluate the effectiveness of these strategies by using a combination of them for detection. Table \ref{tab:diff_analysis} presents the results of these analyses. The results demonstrate that incorporating all the prompting strategies enhances detection performance, leading to improvements in accuracy, precision, and F1 score.

\end{enumerate}


\textbf{Validation Data and Results}:
We conducted validation using two datasets: an internal dataset and the implicit-hate dataset \cite{elsherief-etal-2021-latent}. For implicit-hate, we sampled 250 examples each of Explicit Hate Speech, Implicit Hate Speech, and Normal Speech to ensure a balanced evaluation across different types of speech. In contrast, the internal dataset consisted of 11,645 examples, which, after applying a profanity filter, was reduced to 10,645. Given the unbalanced distribution of hate speech versus normal speech on internal dataset, calculating recall was challenging. As a result, we only focused on the positive predictions generated by each model.

On the internal dataset, the BERT model identified 397 out of 10,645 examples (3.7\%) as positive, while the LLM predicted 509 examples (4.8\%) as positive. To assess these predictions, we randomly sampled 40 examples from each set of positive predictions, which were reviewed by two independent curators, given the subjective nature of the task. While precision varied significantly between curators owing to the subjective nature of the task, the LLM consistently outperformed the BERT model, with an average delta of 7.5\%.

For the implicit-hate dataset, we evaluated various prompt templates and temperature values to enhance the performance of the LLM. A temperature value of 0.6, combined with the prompt template described {\href{https://github.com/AnokiAI/ContextIQ-Paper/blob/master/supplementary/hatespeech_detection/default_prompt_template.yaml}{here}}, yielded the optimal results.
Table \ref{tab:classification_metrics} presents the results for the best parameter combinations for both the LLM-based and BERT models, along with the outcomes for the ensemble models. The ensemble model outperformed the individual models, offering the flexibility to fine-tune precision and recall according to specific requirements. Additionally, the table also provides results for the ensemble model using both AND and OR operations across two different threshold values. The selection of these parameters can be guided by the desired balance between precision and recall in different scenarios.

\end{document}